\documentclass{article}

\usepackage{arxiv}
\usepackage[utf8]{inputenc} 
\usepackage[T1]{fontenc}    
\usepackage{hyperref}       
\usepackage{url}            
\usepackage{booktabs}       
\usepackage{amsfonts}       
\usepackage{nicefrac}       
\usepackage{microtype}      
\usepackage{lipsum}		
\usepackage{graphicx}
\usepackage{natbib}
\usepackage{doi}
\usepackage[utf8]{inputenc}
\usepackage{algorithm}
\usepackage{algpseudocode}
\usepackage{amsmath}

\usepackage{tabularx}
\usepackage[table]{xcolor}
\usepackage{cellspace}
\usepackage{multirow}
\newcolumntype{Y}{>{\centering\arraybackslash}X}
\newcolumntype{Z}{>{\raggedright\arraybackslash}X}

\title{Utilizing GPT to Enhance Text Summarization: A Strategy to Minimize Hallucinations}

\author{
  Hassan Shakil \\
  University of Colorado - Colorado Springs \\
  Colorado Springs, CO \\
  \texttt{hshakil@uccs.edu} \\
  \And
  Zeydy Ortiz \\
  DataCrunch Lab, LLC \\
  Cary, NC \\
  \texttt{zortiz@datacrunchlab.com} \\
  \And
  Grant C. Forbes\\
  North Carolina State University \\
  Raleigh, NC \\     
  \texttt{gforbes@ncsu.edu} \\
}

\date{}

\renewcommand{\shorttitle}{GPT-Enhanced Summarization: Reducing Hallucinations}

\hypersetup{
pdftitle={Utilizing GPT to Enhance Text Summarization: A Strategy to Minimize Hallucinations},
pdfsubject={q-bio.NC, q-bio.QM},
pdfauthor={David S.~Hippocampus, Elias D.~Striatum},
pdfkeywords={First keyword, Second keyword, More},
}

\begin{document}
\maketitle

\begin{abstract}
    In this research, we uses the DistilBERT model to generate extractive summary and the T5 model to generate abstractive summaries. Also, we generate hybrid summaries by combining both DistilBERT and T5 models. Central to our research is the implementation of GPT-based refining process to minimize the common problem of hallucinations that happens in AI-generated summaries. We evaluate unrefined summaries and, after refining, we also assess refined summaries using a range of traditional and novel metrics, demonstrating marked improvements in the accuracy and reliability of the summaries. Results highlight significant improvements in reducing hallucinatory content, thereby increasing the factual integrity of the summaries. \end{abstract}


\section{Introduction}
Robust text summarization technologies are more in demands as digital content spreads quickly over the internet. These technologies makes it easier to consume vast amount information by compressing length texts into brief summaries while maintaining contexts and meanings. Text summarization plays a crucial role in a variety of fields such as corporate reporting, research reviews, and news aggregations by rapidly and effectively delivering pertinent information \cite{antony2023survey}.

Techniques for text summarization is often dividing into two categories; extractive and abstractive methods. In order to provide a logical summary Extractive summarization - which usually makes use of models such as DistilBERT \cite{abdel2022performance} - focuses on identifying and compiling important words or phrases from the original text. On the other side, abstractive summarization - which utilizes models such as T5 \cite{ranganathan2022text} - involves generating new sentences, which frequently results in summaries that are more streamline and succinct. But this approach is especially vulnerable to “hallucinations” - instances where the summary includes plausible but unsupported content from the source text \cite{cao2018faithful}. We also generated hybrid summaries that combines the factual accuracy of extractive summarization with the linguistic complexity of abstractive summarization \cite{el2021automatic}. The technique aim to generate and evaluate a framework for summarization that improves quality while notably lowering hallucinations. In order to further enhance the summaries, our method combines the benefits of extractive and abstractive summarization with a novel refining method using Generative Pre-trained Transformers (GPT) \cite{yenduri2024gpt}.

This study covers the incorporation of a GPT-based refinement process aimed at successfully reducing hallucinatory content. We evaluate this process using an extensive set of metrics, and through rigorous testing and analysis, we demonstrate how our method enhances and refines current summarization methodologies. Our findings aim to show that the generated summaries are not only more accurate and reliable but also concise.

\section{Literature Review}

Text summarization is an important tool for handling the deluge of digital information. Different approaches have been developed, each with its own advantages and disadvantages. Extractive summarization which is known for their simplicity and direct use of text from the source document, have traditionally been favored for their factual accuracy and ease of implementation \cite{allahyari2017text}. In news articles, where the most important information is frequently delivered at the beginning, popular approaches like the Lead-3, which chooses the first few words or sentences of a document \cite{liu2019text}. On the other hand, abstractive summarization generates new sentences, aiming for human-like summaries and has the capacity to paraphrase and generalize from the original material \cite{dreyer2023evaluating}. Even though this method can provide succinct and smooth summaries, it is particularly prone to errors and hallucinations.

The challenge of hallucinations in abstractive summarization has been sparked a lot of research interests, that led to a variety of approach aiming to minimize these errors. While more recent strategies includes the application of advanced machine learning techniques like reinforcement learning, which penalizes the generation of content not found in the source text \cite{narayan2018ranking}, earlier efforts was often leaned on the strict generation constraints or rule-based systems \cite{naik2017extractive}. Furthermore, models explicitly trained to verify the factual accuracy of post-generation summaries. These models measure semantic similarity to ensure the accuracy of the content \cite{reimers2019sentence}.

Hybrid approaches are design for reducing the drawbacks of both extractive and abstractive methods. These techniques combine the coherence and readability of abstractive summarization with the factual correctness of extractive summary. In an effort to generate more accurate and coherent summaries, methods proposed by \cite{zhu2020hierarchical} which includes extract important sentences and then rewriting them using an abstractive mechanism. But even with these advancements, the challenge to effectively manage the balance between accuracy and readability, particularly in minimizing hallucinations, remains a significant concern \cite{ji2023survey}.

In our research, we use the T5 model for abstractive summaries and the DistilBERT model for extractive summaries. We also provide hybrid summaries by the combination of T5 and DistilBERT models. This research primarily focuses on the GPT-based refinement method, which is designed to further minimize hallucinations and improve the summaries' overall reliability. This approach seeks to address the limitations highlighted in prior studies and establish a new benchmark in text summarization technology.

\section{Methodology}

\subsection{Generation of Unrefined Summaries}
\begin{figure}[ht]
\centering
\includegraphics[width=0.75\textwidth]{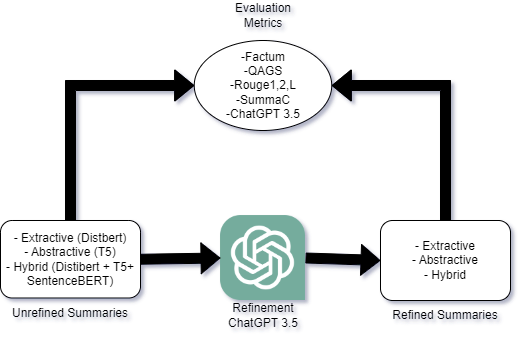}
\caption{Schematic Representation of the Research Methodology}
\label{fig:pipeline}
\end{figure}

   Our methodology begins by generating extractive, abstractive, and hybrid unrefined summaries, each leveraging its distinct strengths to generate comprehensive initial drafts. Figure \ref{fig:pipeline} shows a schematic representation of the research methodology.

\subsubsection{Extractive Summarization Using DistilBERT}
In the first stage of unrefined summary generation, DistilBERT model used to generate extractive summaries. In order to generate a summary that is factually accurate and true to the original content, this method entails picking the most pertinent sentences from the original text \cite{sanh2019distilbert}.

\subsubsection{Abstractive Summarization Using T5}
The T5 model is used to generate abstractive summaries in parallel to extractive summarization. T5 rewrites the text by creating new sentences and distilling the content, seeking for summaries that are not only concise but also have a higher degree of linguistic variety \cite{raffel2020exploring}.

\subsection{Hybrid Summarization Approach}

Post-generation, the outputs from DistilBERT and T5 are funneled into the custom algorithm \ref{alg:hallucination_reduction}. This algorithm is used to enhance the hybrid technique, ensuring a balance between readability and clarity. This refinement systematically filters and retains only those words in the summary that have a high semantic similarity to the words in the original text.

\subsubsection{Algorithm: Hallucination Reduction in Text Summaries}
\begin{algorithm}
\caption{Hallucination Reduction in Text Summaries}
\label{alg:hallucination_reduction}
\begin{algorithmic}[1]
\Require $original\_text$, $summary$
\Ensure $new\_summary$
\State $original\_words \leftarrow \textsc{Tokenize}(original\_text)$
\State $summary\_words \leftarrow \textsc{Tokenize}(summary)$
\State $original\_embeddings \leftarrow \textsc{Encode}(original\_words)$
\State $new\_summary\_words \leftarrow \text{empty list}$
\For{each $word$ in $summary\_words$}
    \State $word\_embedding \leftarrow \textsc{Encode}(word)$
    \State $similarities \leftarrow \textsc{CosineSimilarity}(word\_embedding, original\_embeddings)$
    \If{$\max(similarities) > 0.5$}
        \State \textsc{Append}($new\_summary\_words, word$)
    \EndIf
\EndFor
\State $new\_summary \leftarrow \textsc{Join}(new\_summary\_words)$
\Return $new\_summary$
\end{algorithmic}
\end{algorithm}

\subsection{GPT-Based Evaluation and Refinement Process}
The unrefined (Extractive, Abstractive and Hybrid) summaries are subjected to a refinement process using GPT-based prompts. This stage aims to mitigate any hallucinations and enhance overall summary quality. Each prompt in our process serves to evaluate and refine the summaries by assessing their fidelity and accuracy against the source article.

\begin{algorithm}
\caption{GPT-based evaluation and refinement process}
\begin{algorithmic}[1]
\State \textbf{Step 1: Initial Evaluation}
\State \textit{Prompt for Basic Validation (Yes or No)}
\begin{quote}
"Please determine whether the provided summary can be classified as 'hallucinated' or not by matching it with the appropriate article. Note that, 'hallucination' refers to a summary that is linguistically logical but contains details that are either not mentioned in the article or are factually inaccurate."
\end{quote}
\State \textit{Answer Type:} "Answer (yes or no)"
\State \textbf{Step 2: Detailed Evaluation}
\State \textit{Prompt for Detailed Analysis}
\begin{quote}
"Please determine whether the provided summary can be classified as 'hallucinated' or not by matching it with the appropriate article. Note that, 'hallucination' refers to a summary that is linguistically logical but contains details that are either not mentioned in the article or are factually inaccurate."
\end{quote}
\State \textit{Answer Type:} "Explain your reasoning step by step then answer the question (yes or no)"
\State \textbf{Step 3: Scoring the Summary}
\State \textit{Prompt for Scoring Hallucination Level}
\begin{quote}
"Score the following summary by the given corresponding article with respect to 'hallucination' from 1 to 10. Note that in this context, 'hallucination' refers to a summary that is linguistically logical but contains details that are either not mentioned in the article or are factually inaccurate. 10 points indicates that the summary is not hallucinated at all and 1 point indicates that the summary is fully hallucinated."
\end{quote}
\State \textit{Answer Type:} "Points"
\State \textbf{Step 4: Refinement}
\State \textit{Prompt for Refinement}
\begin{quote}
"Refine the summary and reduce 'hallucination' and try to achieve the score of 10 out of 10 for each summary. Note that in this context, 'hallucination' refers to a summary that is linguistically logical but contains details that are either not mentioned in the article or are factually inaccurate."
\end{quote}
\State \textit{Answer Type:} "Refined Summary"
\State \textbf{Step 5: Final Verification}
\State \textit{Repeat Evaluation for Refined Summaries}
\begin{quote}
"Final Verification: Re-evaluate the refined summaries to confirm the reduction of hallucinations. Repeat the evaluation process for the refined summaries to ensure that they meet the desired criteria of minimal hallucinations. A final round of evaluation confirms the effectiveness of the refinements in reducing hallucinations."
\end{quote}
\State This step involves re-running the initial evaluation prompts on the refined summaries. It seeks to ensure that the modifications have effectively reduced hallucinations and that the final summaries adhere closely to the factual content of the original articles. The outcomes of this final evaluation are documented to provide empirical evidence of the methodology's success.

\end{algorithmic}
\end{algorithm}

\subsection{Implementation Details}

The CNN/Daily Mail dataset \cite{gambhir2022deep} is used for the summarization process, and articles are pre-processed before being processed through the summarization models. The same set of established metrics are used to evaluate both unrefined and refined summaries in order to verify quality improvements and ensure adherence to the integrity of the source material. This consistent evaluation approach allows for a direct comparison of the summaries before and after refinement, highlighting the effectiveness of our refinement process in enhancing factual accuracy and reducing hallucinations. Our approach, which incorporates advanced techniques and rigorous refinement, aim to increase the reliability of text summarization and establish new standards in the field by enhancing the credibility of automated summarizations. This comprehensive approach ensures that our improvements in summary quality are measurable and significant, contributing to the ongoing advancement of summarization technology.

\section{Evaluation Metrics}
A variety of metrics were employed to assess the quality of both the unrefined and refined summaries, with a particular focus on minimizing hallucinations. Our evaluation framework utilizes a blend of conventional and novel metrics to assess the quality and accuracy of the generated summaries comprehensively.

\subsection{FactSumm}
FactSumm utilizes named-entity recognition and relation extraction to extract facts from the source document and the output summary in order to evaluate the consistency of the facts. To assess consistency, this metric compares extracted facts represented as relation triples (subject, relation, object) \cite{factsumm}.

\subsection{QAGS (Asking and Answering Questions)}
QAGS assess the quality of summaries by generating questions from the summary and answering them using the source document. The accuracy of these answers provides a measure of how well the summary represents the factual content of the original text \cite{Wang2020AskingAA}.

\subsection{SummaC}
SummaC employs Natural Language Inference (NLI) approaches to determine if the source document logically implies a summary. It makes use of two models: SummaC-conv, a trained model that aggregates entailment scores using a convolutional layer, and SummaC-zs, which employs zero-shot aggregation of sentence-level scores \cite{10.1162/tacl_a_00453}.

\subsection{ROUGE (Recall-Oriented Understudy for Gisting Evaluation)}
ROUGE metrics examine how the summary and the original document overlap in terms of unigrams (ROUGE-1), bigrams (ROUGE-2), and longest common subsequences (ROUGE-L) \cite{lin-2004-rouge}. ROUGE ratings are a useful tool for assessing textual similarity, although they are not always accurate in terms of facts.

\subsection{Novel Evaluation Using GPT 3.5 Turbo}
\subsubsection{Rationale for Using GPT in Evaluation}
We employ GPT 3.5 Turbo to leverage its advanced language understanding capabilities for evaluating the refined summaries. GPT's ability to understand context and infer logical connections makes it exceptionally suitable for assessing factual consistency and identifying hallucinations.

\subsubsection{Design of Evaluation Metrics with GPT}
The evaluation with GPT is integrated using prompts designed to assess the factual accuracy and detect hallucinations within summaries. These prompts, initially described in the Methodology section under the GPT-based evaluation and refinement process, serve multiple roles:
\begin{itemize}
    \item \textbf{Scoring Hallucinations:} We reuse the prompt from the final refinement step in the methodology to score the summaries on a scale of hallucination. This prompt assesses the level of hallucination from 0 to 1, where 0 indicates a fully hallucinated summary and 1 indicates no hallucination.
    \item \textbf{Detailed Analysis:} The same detailed analysis prompt used in the methodology section is employed here to explain the reasoning behind the scores given, ensuring a comprehensive evaluation.
\end{itemize}

\subsubsection{Comparison with Traditional Evaluation Metrics}
Unlike traditional metrics such as ROUGE, which primarily measure textual overlap, the GPT-based evaluation provides a deeper insight into the semantic and factual correctness of the summaries. This approach ensures that the summaries not only share lexical similarities with the source texts but also adhere closely to factual accuracy, addressing the key concern of hallucination more effectively.

\section{Results}

\begin{figure}[ht]
\centering
\includegraphics[width=0.75\textwidth]{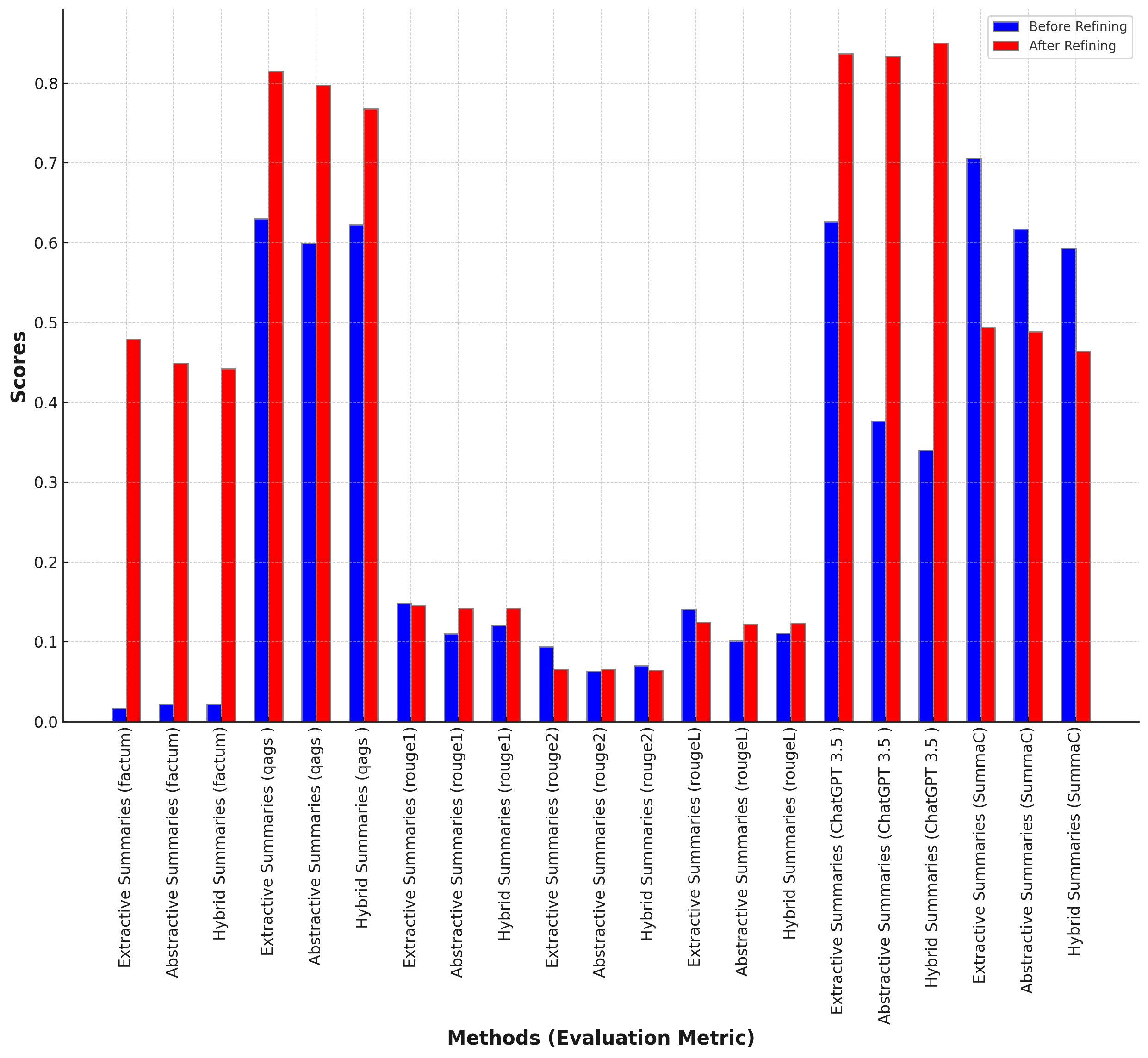}
\caption{Comparison of evaluation metrics for unrefined and refined summaries.}
\label{fig:eval_metrics}
\end{figure}

Our analysis was structured to evaluate the performance of unrefined and refined summaries generated by extractive (DistilBERT), abstractive (T5), and our hybrid approach. The evaluation metrics, including FactSumm, QAGS, GPT 3.5, SummaC, and ROUGE, provided a multifaceted view of the summaries' quality, depicted in the bar chart (see Figure \ref{fig:eval_metrics}).

\subsection{Improvements in Factual Consistency and Hallucination Reduction}
The QAGS and FactSumm evaluations demonstrated (see Figure \ref{fig:eval_metrics}) an increase in scores after refinement across all types of summaries, with the most pronounced improvement observed in abstractive summaries. This suggests that the refinement process effectively reduced hallucinated content and enhanced factual consistency.

The evaluation using GPT 3.5 revealed a significant score improvement in abstractive and hybrid summaries post-refinement. This advancement can be attributed to the model's ability to grasp nuanced semantic relationships and identify factually inconsistent information that was corrected during the refinement process.

\subsection{Varied Responses from SummaC and ROUGE Metrics}
Surprisingly, after refinement, SummaC scores decreased, especially for abstractive summaries. This might be because Natural Language Inference (NLI) is used by the SummaC metrics to assess the logical entailment between the summary and the original text. Lower scores may have resulted from our refinement process' inclusion of more intricate sentence structures or paraphrases that do not align with NLI patterns that SummaC expects. This indicates an area for potential improvement in our refinement process to better align with NLI-based evaluation methods.

In terms of the ROUGE metrics, scores for extractive summaries showed improvement in ROUGE-1 and ROUGE-L, whereas scores for abstractive and hybrid summaries showed variation. ROUGE-2 scores did not show a clear trend of improvement. The quality of semantic relationships or factual accuracy may not be effectively captured by the ROUGE suite, which primarily assesses overlap in n-grams between the summary and source text. Therefore, abstractive summaries with more paraphrased information did not consistently reflect higher ROUGE scores post-refinement, even though extractive summaries—which are more closely linked with the original text—showed improvement.




\section{Statistical Analysis}

This section evaluates the efficacy of our text summarization refinement process through statistical methods, aiming to determine if observed improvements are statistically significant and not due to random variation. We conducted paired t-tests to compare evaluation metric scores before and after the refinement process, appropriate for analyzing two related samples. This method assesses whether the mean difference between paired observations (pre- and post-refinement scores) is statistically significant \cite{field2013discovering}.

\subsection{Hypothesis Formulation}
The null hypothesis $(H_0)$ for our tests posited that the mean score of the refined summaries would not be greater than the mean score of the unrefined summaries. The alternative hypothesis $(H_1)$ suggested that the refined summaries would have a higher mean score \cite{lehmann2005testing}.

\subsection{Results Interpretation}
Statistical analysis, as illustrated in Table \ref{tab:results}, indicated significant improvements in several metrics post-refinement $(p < 0.05)$, leading to the rejection of the null hypothesis for metrics such as FactSumm, QAGS, GPT 3.5, ROUGE-1, and ROUGE-L. These metrics showed substantial mean score increases, indicated by their negative values in the 95\% Confidence Interval column, suggesting effective refinement.

Conversely, ROUGE-2 and SummaC did not achieve statistical significance, with p-values above 0.05, indicating no substantial improvement after refinement for these metrics.

\subsection{Discussion and Implications}
The statistical analysis indicates that our ``Refine Prompt" approach has generally improved the quality of summaries according to most metrics. However, the variability in the results, especially for ROUGE-2 and SummaC, invites further scrutiny.

The decline in SummaC scores post-refinement for abstractive summaries, coupled with the inconsistency in ROUGE-2 scores, suggests that the refinement process may affect different types of summaries in different ways. This could be due to various factors, such as the introduction of complex language structures during refinement that do not align with the patterns recognized by these metrics as indicative of high quality. It raises important considerations for future refinements of the summarization algorithms, particularly to ensure alignment with NLI-based evaluation methods and lexical overlap metrics.

The findings underscore the need for a comprehensive set of evaluation metrics that can capture the multifaceted nature of summary quality, going beyond mere word overlap to encompass semantic and factual integrity.

\begin{table}[h]
\centering
\caption{Summary of Hypothesis Testing Results}
\label{tab:results}
\begin{tabularx}{\textwidth}{|Z|Y|Y|Y|Y|Y|}
\hline
\rowcolor{gray!30}
\textbf{Evaluation Metrics} & \textbf{Average Before Refining} & \textbf{Average After Refining} & \textbf{p-value} & \textbf{95\% Confidence Interval} & \textbf{Reject Null?} \\
\hline
FactSumm & 0.02 & 0.46 & $<$0.001 & (-0.53, -0.35) & Yes \\
\hline
QAGS & 0.62 & 0.77 & $<$0.001 & (-0.23, -0.12) & Yes \\
\hline
Rouge1 & 0.12 & 0.14 & $<$0.001 & (-0.02, -0.01) & Yes \\
\hline
Rouge2 & 0.07 & 0.06 & $\approx$ 1.0 & (0.004, 0.01) & No \\
\hline
RougeL & 0.11 & 0.12 & 0.031 & (-0.01, 0.0003) & Yes \\
\hline
SummaC & 0.64 & 0.48 & $\approx$ 1.0 & (0.10, 0.21) & No \\
\hline
GPT 3.5 & 0.45 & 0.84 & $<$0.001 & (-0.45, -0.34) & Yes \\
\hline
\end{tabularx}
\end{table}

\subsection{Correlation Analysis}
\begin{figure}[ht]
\centering
\includegraphics[width=0.75\textwidth]{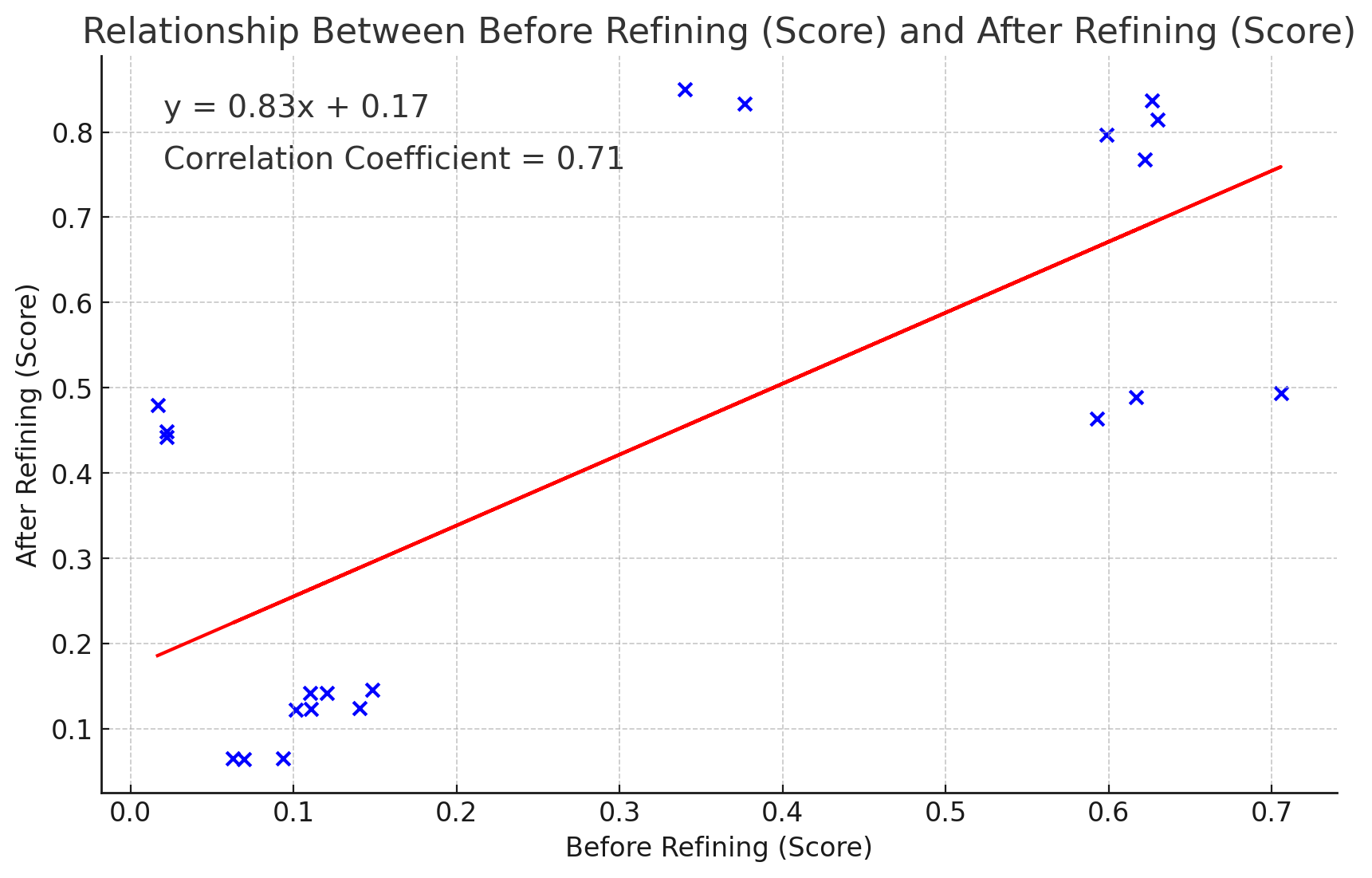}
\caption{Scatter plot illustrating the correlation between pre- and post-refinement scores, with the line of best fit highlighting the general trend of improvement.}
\label{fig:scatterplot}
\end{figure}

To substantiate our hypothesis testing, we conducted a correlation analysis on the scores pre- and post-refinement. The strong positive correlation coefficient, recorded at 0.71, implies that summaries with higher pre-refinement scores tended to exhibit more pronounced improvements post-refinement. This trend is clearly illustrated in the scatter plot (Figure \ref{fig:scatterplot}), where the data points are effectively captured by the regression line, depicted by the equation y = 0.83x + 0.17, which underscores the positive trajectory of score enhancement through the refinement process.
   
\section{Discussion}

The analysis of our evaluation data reveals crucial insights into the performance of the summarization refinement process. While the FactSumm and QAGS metrics show clear improvements post-refinement, the decline in SummaC scores for abstractive summaries suggests a complex interplay between the factual alignment and the semantic structure recognized by NLI-based metrics. The variations in ROUGE scores highlight a similar complexity, implying that lexical overlap does not always correlate with improved summary quality.

These results prompt a discussion on the adequacy of current evaluation metrics for summarization. Metrics that rely heavily on lexical overlap may not adequately convey the essence of what makes an improved summary, particularly when advanced language models are used for refinement. This emphasizes the need for further research to create more sophisticated evaluation frameworks that are better able to recognize the nuances of language model-based summary refinement.

The results also highlight how important context and factual consistency are when evaluating summary quality, a area in which large language models like GPT perform exceptionally well. As summarization technology evolves, the need for evaluation methods that keep pace with these advancements becomes increasingly apparent.

In conclusion, our results affirm the potential of advanced language models in improving text summarization, but also highlight the need for a more nuanced approach to their evaluation. Future work will focus on refining these methods further and exploring evaluation metrics that can more accurately reflect the quality of generated summaries.

\section{Conclusions}
We have generated extractive, abstractive, and hybrid summaries, which we subsequently refined using an innovative GPT-based process. This novel approach in our methodology demonstrates promise in overcoming traditional summarization constraints, notably in reducing hallucinations and improving factual consistency. Our evaluations show significant improvements across several metrics, substantiating the effectiveness of our refinement in bolstering summary quality. However, the varied responses from metrics like SummaC and ROUGE-2 highlight the ongoing challenges in summary evaluation, suggesting a need for evaluation frameworks that can better accommodate the complexities introduced by advanced language models. In the future, we want to improve these evaluation methods to offer more accurate assessments of summary quality, ensuring that our advancements in summarization technology can be fully realized and appropriately validated.


\bibliographystyle{unsrtnat}
\bibliography{references}  

\end{document}